# GENERATION OF SYNTHETIC DATA USING BREAST CANCER DATASET AND CLASSIFICATION WITH RESNET18

***Research Article***

Dilşat Berin AYTAR[1], Semra GÜNDÜÇ[2]

[1]*Ankara University, Department of Computer Engineering, Ankara, Turkey, aytar.berin@gmail.com*
[2]*Ankara University, Department of Computer Engineering, Ankara, Turkey, Semra.Gunduc@ankara.edu.tr*

## ABSTRACT

Since technology is advancing so quickly in the modern era of information, data is becoming an essential resource in many fields. Correct data collection, organization, and analysis make it a potent tool for successful decision-making, process improvement, and success across a wide range of sectors. Synthetic data is required for a number of reasons, including the constraints of real data, the expense of collecting labeled data, and privacy and security problems in specific situations and domains. For a variety of reasons, including security, ethics, legal restrictions, sensitivity and privacy issues, and ethics, synthetic data is a valuable tool, particularly in the health sector. A deep learning model called GAN (Generative Adversarial Networks) has been developed with the intention of generating synthetic data.

In this study, the Breast Histopathology dataset was used to generate malignant and negatively labeled synthetic patch images using MSG-GAN (Multi-Scale Gradients for Generative Adversarial Networks), a form of GAN, to aid in cancer identification. After that, the ResNet18 model was used to classify both synthetic and real data via Transfer Learning. Following the investigation, an attempt was made to ascertain whether the synthetic images behaved like the real data or if they are comparable to the original data.

**Keywords—Generative Adversarial Networks, MSG-GAN, gradient, histopathology, CNN, classification, Transfer Learning, ResNet18, breast cancer, synthetic data.**

## 1. INTRODUCTION

In today's world, data has become an extremely important resource in many fields, akin to a valuable commodity. The usability of data in numerous areas has led to the emergence of new data and an increased need for more data. Fields such as scientific research, decision-making processes, strategic planning, performance monitoring, Artificial Intelligence (AI) and Machine Learning (ML), customer experience and personalization, scientific research and innovation, healthcare services and medicine, security and risk management, solving societal issues, and marketing demonstrate the significance of data.

Data plays many important roles in the field of healthcare and medicine, contributing significantly to the development of healthcare services, patient treatment, and the formation of health policies. However, there are limitations to accessing real data in healthcare. Health data often contains sensitive and confidential information. The sharing and access of medical data are subject to strict regulations and can be limited due to privacy concerns, ethics, and security. Examples of legal regulations governing health data include HIPAA (Health Insurance Portability and Accountability Act) in the United States, GDPR (General Data Protection Regulation) in the European Union, and various other regulations in different countries related to the protection and processing of health data. Furthermore, health data is protected under medical ethical rules and standards. These principles address issues such as data confidentiality, patient privacy, and patient rights, guiding healthcare providers. Therefore, healthcare institutions and providers must strictly adhere to these regulations and take various measures to ensure the confidentiality and security of health data, ensuring patient safety and preventing data misuse. Overcoming limitations in accessing data is crucial for health research and innovations.

Overcoming limitations and addressing data scarcity, synthetic data generated by artificial intelligence presents a significant solution. Synthetic data, created artificially by a computer program, is designed to mimic the characteristics of real-world data while preserving individual privacy and avoiding data breaches. Organizations can generate nearly unlimited amounts of data for testing, research, and analysis using synthetic data without worrying about ethical and legal issues associated with real-world data. Generative Adversarial Networks (GANs) have emerged for the purpose of synthetic data generation.

GANs are a significant and innovative modeling approach in the field of deep learning. They consist of two networks, a generator and a discriminator, which compete with each other. These networks compete to ultimately produce realistic images. Various types of GAN model variants have been developed to meet different needs. This article will describe Multi-Scale Gradients for Generative Adversarial Networks (MSG-GAN), a type of GAN model.

In MSG-GAN, the generator and discriminator networks compete at a single resolution and improve together. The Multi-Scale Gradients technique utilizes different resolution levels to stabilize this competition. This approach gradually increases operations starting from lower resolutions and scales up to real dimensions. As a result, a faster, more stable, and improved training process is provided, contributing to more realistic, consistent, and high-quality results by better utilizing information at different scales. MSG-GAN is particularly successful in tasks such as image generation and synthesis involving visual datasets. It has a wide range of applications in data augmentation, artificial intelligence studies, art, computing, medicine, automotive, finance, and many other fields.

After synthetic data generation with MSG-GAN, the generated synthetic data will be classified using Transfer Learning techniques such as ResNet18 (Residual Neural Network). ResNet18 is a deep learning model commonly used for visual recognition and classification problems. The term "ResNet" stands for "Residual Networks," and "18" denotes the number of layers in the model.

ResNet represents an architecture that includes residual blocks developed to facilitate the training of deep neural networks and reduce overfitting. The residual block passes the input data through an activation function and several convolutional and summation layers. An important feature of ResNet is the presence of "residual connections" in these blocks.

Transfer Learning refers to the reuse of features learned by a pre-trained model to solve a different task. If necessary, weights are adjusted and new layers are added based on the new task and dataset. For example, a pre-trained ResNet18 model may have been trained for many image classification tasks. In a new image classification task, the pre-trained ResNet18 model can be taken and retrained on the new dataset.

Breast cancer is the second most common cancer type globally after lung cancer (Teh et al., 2015). Invasive Ductal Carcinoma (IDC) is the most common subtype among all breast cancers. The aim is to reduce reliance on pathologists and thereby reduce errors and human-related biases during disease detection, as well as minimize the high economic cost and time loss associated with it. In this study, IDC+ and IDC- histopathological images will be generated using MSG-GAN for disease detection, and the images will be classified using ResNet18.

## 2. RELATED WORK

In the literature, several studies have been conducted on the generation and classification of synthetic data using GANs from breast cancer datasets.

Wu et al. (2018) developed Mammo-ciGAN using the DDSM dataset to augment Film Mammography (MMG) data and add synthetic lesions to healthy mammograms. The results obtained using Mammo-ciGAN were evaluated using the ROC AUC (Receiver Operating Characteristic- Area Under the Curve) metric. The ROC AUC value of 88.7% reflects the potential of using the model to add synthetic lesions to healthy mammograms.

Guan and Loew (2019) aimed to augment Film Mammography (MMG) data using GAN with the DDSM dataset. According to the results, the accuracy rate of film mammography data augmentation using this GAN model is 73.48%. This augmentation process focused on creating patches containing both benign and malignant tumors.

Desai et al. (2020) synthetically generated Film Mammography (MMG) images using the DDSM dataset and then used these images in a visual Turing test. The results show that the accuracy rate of the synthetically generated images is 78.23%.

Kansal et al. (2020) worked on a specific Optical Coherence Tomography (OCT) dataset using DCGAN. In this study, data augmentation was aimed at generating synthetic OCT images. According to the results, the accuracy rate of generating synthetic OCT images is 92%.

Shahidi (2021) aimed to generate high-resolution synthetic images from low-resolution images using WA-SRGAN with a combination of breast and lymph node histopathology datasets. The study achieved a 99.49% accuracy value in high-resolution images. The use of higher-resolution images can improve diagnostic accuracy and enable more precise detections.

# 3. METHODOLOGY

In the two-stage study, in the first stage, synthetic images were produced using MSG-GAN, and in the second stage, classification was made using ResNet18, one of the Transfer Learning techniques. Finally, the classification results were evaluated with metrics.

## 3.1 MSG-GAN (Multi-Scale Gradients Generative Adversarial Network)

MSG-GAN (Karnewar et al., 2020), is a technique used to enhance the performance of traditional GANs (Goodfellow et al., 2014) by stabilizing their training process and achieving high-quality results.

The Multi-Scale Gradients technique uses different resolution levels (starting from lower to higher resolutions) to stabilize this competition. This approach begins with lower resolutions and gradually scales up operations to the actual size. This ensures a faster, more stable, and improved training process, leveraging information at different scales to produce more realistic, consistent, and high-quality results.

The process generally involves the following steps:

1. **Start at Low Resolution**: Initially, the generator network operates at a lower resolution, learning simpler patterns.
2. **Increase Resolution**: As the generated images are scaled to higher resolutions, the complexity of the network increases, adding more detail and realism.
3. **Finalize at Real Size**: The process continues until the target resolution is reached, allowing the network to learn more complex patterns and details.

This technique is considered a significant advancement in the evolution of GANs, often leading to better performance on high-resolution images or other complex data types. MSG-GAN is utilized in various image synthesis and other application areas.

Figure 1 shows the basic MSG-GAN architecture used in the study. The architecture includes connections from the intermediate layers of the generator to the intermediate layers of the discriminator. The multi-scale images sent to the discriminator are combined with activation volumes obtained from the main path of the convolutional layers using a "Combine Function" (shown in yellow) (Karnewar et al., 2020).

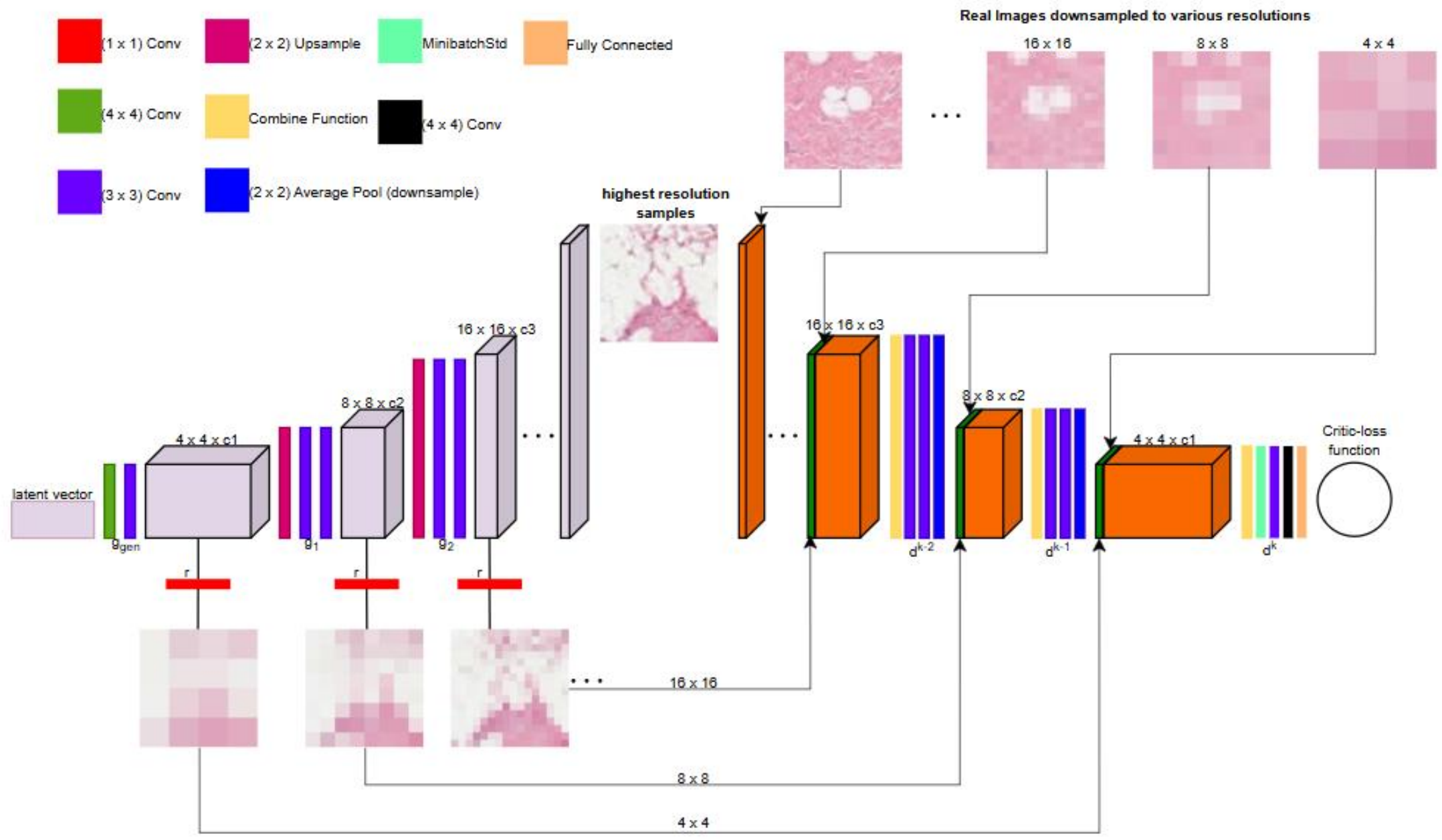

Figure 1. MSG-GAN architecture

### 3.1.1 MSG-GAN Generator Architecture and Function

The generator architecture used to produce a 64x64x3 image with MSG-GAN typically consists of 5 blocks. While Table 1, illustrates the entire generator architecture, the 5 blocks used to generate 64x64x3 images in the study are highlighted in bold.

The generator architecture generally involves upsampling and convolutional operations. Upsampling refers to transforming images into higher resolutions. After each upsampling step, two convolution operations are usually performed. The generator processes input noise to generate realistic images with dimensions of 64x64x3.

The generator typically takes random noise vectors as input. This noise is processed and transformed into a feature map containing pixel values. After upsampling the images, the generator typically learns features using convolutional layers with 3x3 filters. Higher-level features are learned at each layer. Initially, these features may represent simple patterns and shapes, which later evolve into more complex objects and structures. The LeakyReLU activation function (Xu et al., 2015) is used in each block.

The generator performs two transmission operations in each block:

- Firstly, after every second convolution operation, the generator passes its output to the next block until reaching the final block. The output size, for example, is 512x16x16.
- Secondly, in addition to Table 1, a 1x1 convolution operation is applied to the output of the respective block before transitioning to the next block. This ensures that the block output has 3 channels (RGB features). This output is transmitted as input to the discriminator. The output size, for example, is 3x16x16.

By following the specified steps and performing the 1x1 convolution operation, the generator produces high-quality images of size 64x64x3 in five blocks and transmits them to the discriminator.

The generator is trained using feedback from the discriminator. The discriminator evaluates the realism of the generated images and provides feedback. The generator learns to make the generated images increasingly realistic based on this feedback. During the training process, the generated images are optimized to resemble real images as closely as possible. This helps prevent issues such as mode collapse and training instability.

Table 1. Generator architecture

| Block | Operation | Activation Function | Output Shape |
|---|---|---|---|
| **1.** | **Latent vector**<br>**Conv 4x4**<br>**Conv 3x3** | **Norm**<br>**LReLU**<br>**LReLU** | **512x1x1**<br>**512x4x4**<br>**512x4x4** |
| **2.** | **Upsample**<br>**Conv 3x3**<br>**Conv 3x3** | **-**<br>**LReLU**<br>**LReLU** | **512x8x8**<br>**512x8x8**<br>**512x8x8** |
| **3.** | **Upsample**<br>**Conv 3x3**<br>**Conv 3x3** | **-**<br>**LReLU**<br>**LReLU** | **512x16x16**<br>**512x16x16**<br>**512x16x16** |
| **4.** | **Upsample**<br>**Conv 3x3**<br>**Conv 3x3** | **-**<br>**LReLU**<br>**LReLU** | **512x32x32**<br>**512x32x32**<br>**512x32x32** |
| **Model 1 ↑** | | | |
| **5.** | **Upsample**<br>**Conv 3x3**<br>**Conv 3x3** | **-**<br>**LReLU**<br>**LReLU** | **512x64x64**<br>**256x64x64**<br>**256x64x64** |
| 6. | Upsample<br>Conv 3x3<br>Conv 3x3 | -<br>LReLU<br>LReLU | 256x128x128<br>128x128x128<br>128x128x128 |
| Model 2 ↑ | | | |
| 7. | Upsample<br>Conv 3x3 | -<br>LReLU | 128x256x256<br>64x256x256 |
| | Conv 3x3 | LReLU | 64x256x256 |
| Model 3 ↑ | | | |
| 8. | Upsample<br>Conv 3x3<br>Conv 3x3 | -<br>LReLU<br>LReLU | 64x512x512<br>32x512x512<br>32x512x512 |
| 9. | Upsample<br>Conv 3x3<br>Conv 3x3 | -<br>LReLU<br>LReLU | 32x1024x1024<br>16x1024x1024<br>16x1024x1024 |
| Model full ↑ | | | |

### 3.1.2. MSG-GAN Discriminator Architecture and Function

The discriminator architecture used to generate a 64x64x3 image with MSG-GAN typically consists of 5 blocks. Table 2 displays the entire discriminator architecture, with the 5 blocks highlighted in bold for generating the 64x64x3 images in the study. The 5 blocks used in the discriminator architecture are the last 5 blocks compared to the first 5 blocks used in the generator architecture. This is because, as described below, the output of each block from the generator will be the input to the corresponding blocks of the discriminator from end to start.

The discriminator model is structured to handle images of different sizes in each block. Each block in Table 2 represents a different scale level. Block operations typically involve taking the raw RGB image, concatenating it with feature maps from previous blocks (Concat/ϕ_simple), adding minibatch standard deviation (MiniBatchStd) to the feature maps, applying a 3x3 convolution operation (a 3x4 convolution operation is applied in the last block), and performing average pooling (Avg Pooling). Each convolution layer uses a certain number of filters, and the LeakyReLU activation function (Xu et al., 2015) is used.

In the last block, average pooling is not performed. Instead, there is a fully connected layer. The fully connected layer produces an output to determine whether an input image of size 64x64x3 is real or fake. During training, it attempts to produce high output values for real images and low output values for generated images as much as possible.

Table 2. Discriminator architecture

| Block | Operation | Activation Function | Output Shape |
|---|---|---|---|
| Model full ↓ | | | |
| | Raw RGB images 0 | - | 3x1024x1024 |
| 1. | FromRGB 0<br>MiniBatchStd<br>Conv 3x3<br>Conv 3x3<br>AvgPool | -<br>-<br>LReLU<br>LReLU<br>- | 16x1024x1024<br>17x1024x1024<br>16x1024x1024<br>32x1024x1024<br>32x512x512 |
| 2. | Raw RGB images 1<br>Concat/$\phi_{simple}$<br>MiniBatchStd<br>Conv 3x3<br>Conv 3x3<br>AvgPool | -<br>-<br>-<br>LReLU<br>LReLU<br>- | 3x512x512<br>35x512x512<br>36x512x512<br>32x512x512<br>64x512x512<br>64x256x256 |
| Model 3 ↓ | | | |
| 3. | Raw RGB images 2<br>Concat/$\phi_{simple}$<br>MiniBatchStd<br>Conv 3x3<br>Conv 3x3<br>AvgPool | -<br>-<br>-<br>LReLU<br>LReLU<br>- | 3x256x256<br>67x256x256<br>68x256x256<br>64x256x256<br>128x256x256<br>128x128x128 |
| Model 2 ↓ | | | |
| 4. | Raw RGB images 3<br>Concat/$\phi_{simple}$<br>MiniBatchStd<br>Conv 3x3<br>Conv 3x3<br>AvgPool | -<br>-<br>-<br>LReLU<br>LReLU<br>- | 3x128x128<br>131x128x128<br>132x128x128<br>128x128x128<br>256x128x128<br>256x64x64 |
| **5.** | **Raw RGB images 4**<br>**Concat/$\phi_{simple}$**<br>**MiniBatchStd**<br>**Conv 3x3**<br>**Conv 3x3**<br>**AvgPool** | **-**<br>**-**<br>**-**<br>**LReLU**<br>**LReLU**<br>**-** | **3x64x64**<br>**259x64x64**<br>**260x64x64**<br>**256x64x64**<br>**512x64x64**<br>**512x32x32** |
| **Model 1 ↓** | | | |
| **6.** | **Raw RGB images 5**<br>**Concat/$\phi_{simple}$**<br>**MiniBatchStd**<br>**Conv 3x3**<br>**Conv 3x3**<br>**AvgPool** | **-**<br>**-**<br>**-**<br>**LReLU**<br>**LReLU**<br>**-** | **3x32x32**<br>**515x32x32**<br>**516x32x32**<br>**512x32x32**<br>**512x32x32**<br>**512x16x16** |
| **7.** | **Raw RGB images 6**<br>**Concat/$\phi_{simple}$**<br>**MiniBatchStd**<br>**Conv 3x3**<br>**Conv 3x3**<br>**AvgPool** | **-**<br>**-**<br>**-**<br>**LReLU**<br>**LReLU**<br>**-** | **3x16x16**<br>**515x16x16**<br>**516x16x16**<br>**512x16x16**<br>**512x16x16**<br>**512x8x8** |
| **8.** | **Raw RGB images 7**<br>**Concat/$\phi_{simple}$**<br>**MiniBatchStd**<br>**Conv 3x3**<br>**Conv 3x3**<br>**AvgPool** | **-**<br>**-**<br>**-**<br>**LReLU**<br>**LReLU**<br>**-** | **3x8x8**<br>**515x8x8**<br>**516x8x8**<br>**512x8x8**<br>**512x8x8**<br>**512x4x4** |
| **9.** | **Raw RGB images 8**<br>**Concat/$\phi_{simple}$**<br>**MiniBatchStd**<br>**Conv 3x3**<br>**Conv 3x4**<br>**Fully Connected** | **-**<br>**-**<br>**-**<br>**LReLU**<br>**LReLU**<br>**Linear** | **3x4x4**<br>**515x4x4**<br>**516x4x4**<br>**512x4x4**<br>**512x1x1**<br>**1x1x1** |

### 3.2 ResNet18

ResNet (Residual Neural Network) (He et al., 2015) is a neural network model introduced to facilitate the training of deep networks and increase performance in the field of visual processing. The main purpose of ResNet is to reduce training difficulties that may occur by making the network deeper.

ResNet uses the concept of residual learning. In this approach, the network tries to learn the difference between the input data and the output, instead of just predicting the outputs of the layers. The basic structural unit of ResNet is the Residual Block. This block is slightly different from a traditional neural network layer. A Residual Block consists of convolution, activation, normalization layers and Residual Connection (Shortcut). Residual connections, which ease the training of deep neural networks, enhance their performance, and enable deeper networks.

ResNet is named according to the number of layers used. In this study, the ResNet18 model with 18 layers was used. ResNet18, a pre-trained Transfer Learning model, is used to perform binary classification in this study.

ResNet18 comprises convolutional layers, batch normalization, and ReLU activation functions (Liu, 2017), Residual connections involve adding the output of the previous layer to the next layer. This technique mitigates the vanishing gradient problem and allows the training of very deep networks possible.

ResNet18 and other ResNet architectures are applied in many areas such as image processing, object recognition, face recognition and medical image analysis. In particular, Transfer Learning techniques and pre-trained networks such as ResNet18 are used to achieve high success rates in various tasks.

In the study, all layers of the pre-trained model except the last layer are frozen and will not be updated during training. Only the last Full Connection layer of the model was changed and the output was arranged to be two classes (positive or negative). The parameters (weights and bias) of the last added Full Connectivity layer are set to be open to training. Figure 2 shows the ResNet18 architecture used in the study.

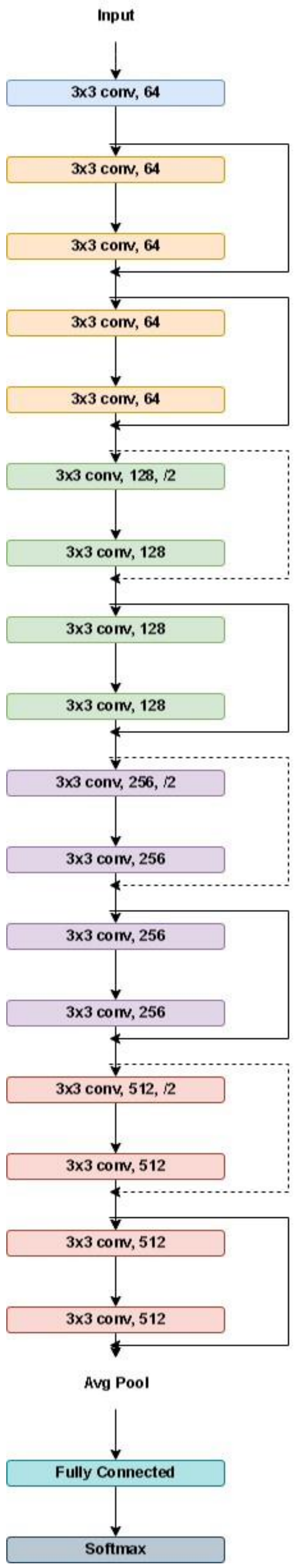

Figure 2. ResNet18 architecture

## 4. EXPERIMENTS AND RESULTS

A certain number of datasets were divided into two parts, one used to generate breast cancer negative and positive labeled images using MSG-GAN. The generated images and the unused portion of the real data were used as training and test data, and four different classification processes were performed using ResNet18. The model was evaluated based on the results obtained.

## 4.1 Dataset and Summary of Work

Firstly, the training dataset was downloaded from https://www.kaggle.com/datasets/paultimothymooney/breast-histopathology-images. Janowczyk et al., 2016; Cruz-Roa et al., 2014). Patches of Invasive Ductal Carcinoma (IDC), the most prevalent subtype of breast cancer, are seen in the dataset. Of all breast cancers, invasive ductal carcinoma (IDC) is the most prevalent subtype. Pathologists normally concentrate on areas with IDC when determining the aggressiveness levels of each assembly sample. Thus, locating the precise IDC zones across the entire assembly slide is a typical preprocessing step for automatic aggressiveness rating. There are 156,000 patches in all, consisting of 78,000 IDC negative and 78,000 IDC positive patches. The pictures have three channels and a 50x50 dimension. Data that is IDC negative is labeled 0, whereas data that is IDC positive is labeled 1.

The dataset was divided into two parts for synthetic data generation (40,000 x 2) and classification (38,000 x 2). The reason for this is to ensure that the test data consists of previously unseen data during the training phase when performing classification.

The study consists of two parts.

- Firstly, synthetic data generation was performed using MSG-GAN with 80,000 images. As a result of this study, 38,000 synthetic IDC negative and 38,000 synthetic IDC positive data were generated.
- Secondly, four different classifications were performed using ResNet18, which is one of the Transfer Learning models.

## 4.2 Synthetic Data Generation with MSG-GAN

As explained in section 3.1.1 and section 3.1.2, a 5-block generator and discriminator architecture was used with MSG-GAN. In the study, WGAN-GP (Gulrajani et al., 2017) was determined as the loss function and RMSprop (Graves, 2013) was determined as the optimization algorithm for both the generator and discriminator models. As a result, 38,000 synthetic IDC- and 38,000 synthetic IDC+ data of size 64x64x3 were produced by using 40,000 IDC+ and 40,000 IDC- data from the real data set.

Figure 3 shows examples of synthetic IDC+ data produced using MSG-GAN, and figure 4 shows examples of synthetic IDC- data produced using MSG-GAN.

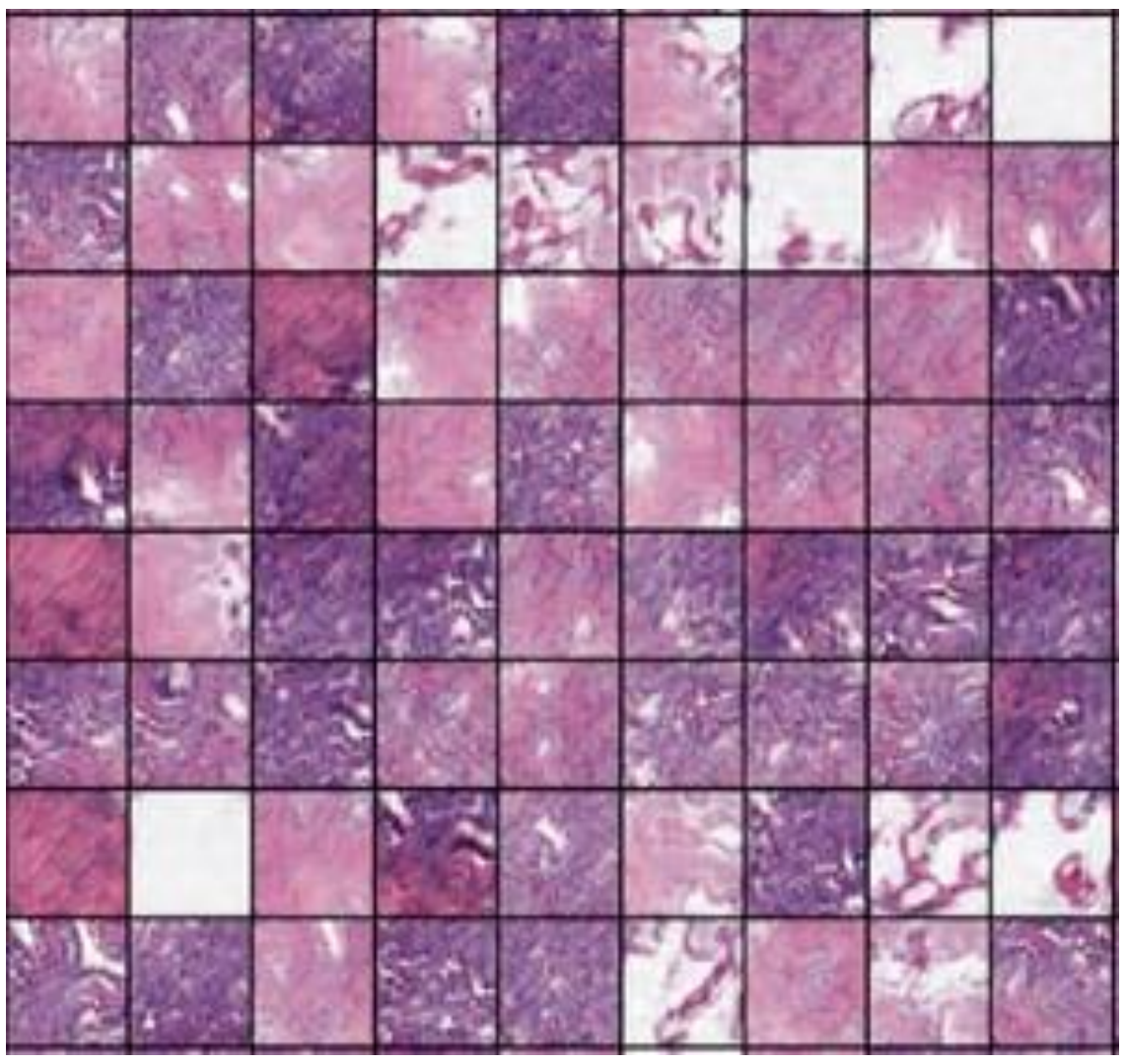

Figure 3. Synthetic IDC+ data examples

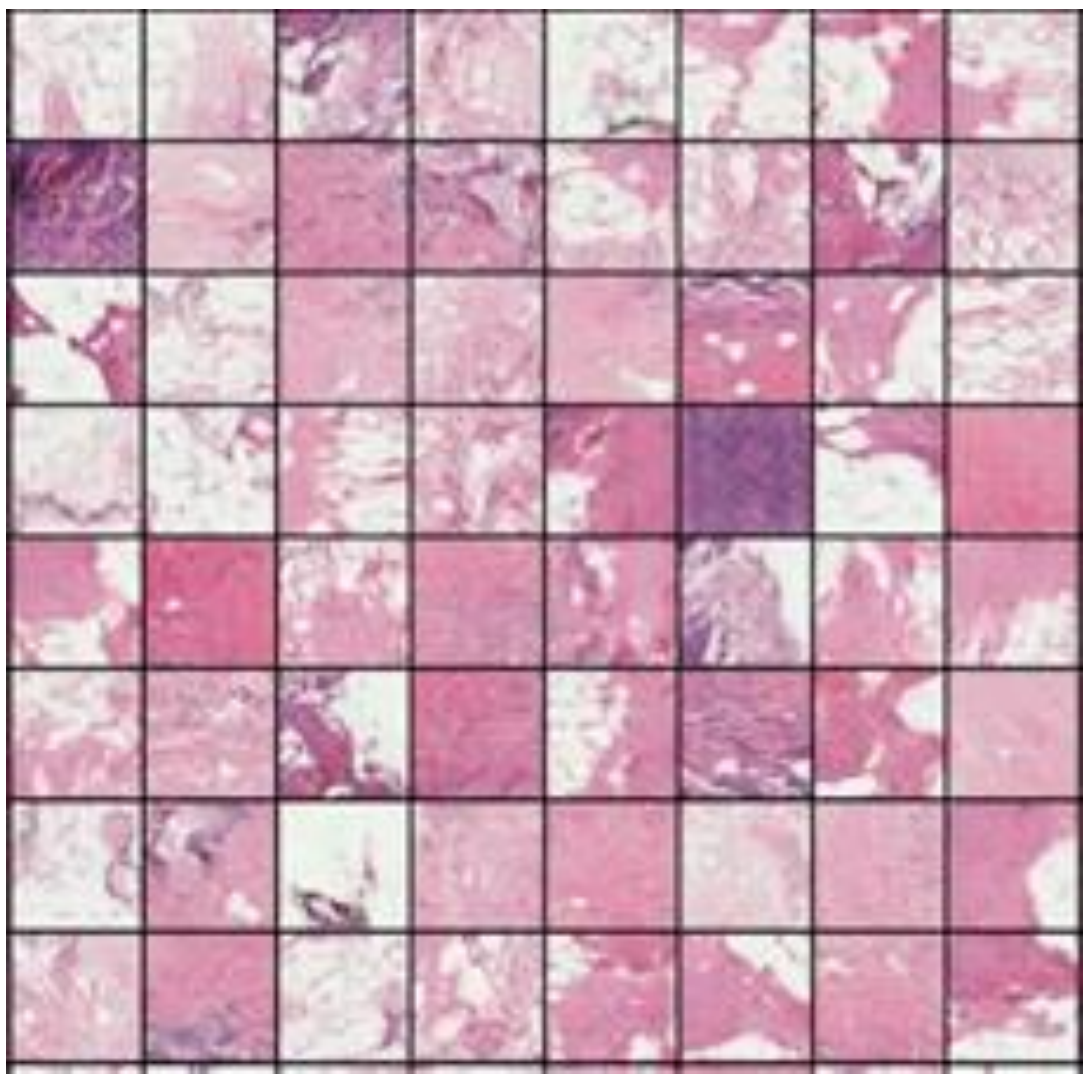

Figure 4. Synthetic IDC- data examples

38,000 synthetic IDC+ and 38,000 synthetic IDC- data produced using MSG-GAN were used in the next stage of the study, namely classification.

### 4.3 Classification of Real and Synthetic Data with ResNet18

In this stage, classification was performed using the unused 76,000 portion of the real dataset and the generated 76,000 synthetic data with the pre-trained ResNet18 model. The purpose of this section was to determine how similar synthetic data was generated to the real data and to classify images. This section was divided into four sub-sections. The same model was used in each subsection. The subsections are as follows:

• Training and classification of real data: 70% (53,200 patches) of the real data was used as training and 30% (22,800 patches) as testing.

• Training and classification of synthetic data: 70% (53,200 patches) of the synthetic data was used as training and 30% (22,800 patches) as testing.

• Training with real data and classification of synthetic data: All real data (76,000 patches) was used as training and all synthetic data (76,000 patches) was used as test data.

• Training with synthetic data and classification of real data: All synthetic data (76,000

patches) was used as training data and all real data (76,000 patches) was used as test data.

A deep learning model like ResNet18 generally expects an input size of 224x224x3. Therefore, in the study, real and synthetic data were resized to 224x224x3 dimensions before the classification process. The ResNet18 model architecture described in Section 3.2 was implemented.

CrossEntropyLoss (LeCun et al., 2018) was used as the loss function, and Adam (Kingma and Ba, 2015) was used as the optimizer.

Training lasted for 150 epochs for each classification. Loss was calculated for each batch, gradients were propagated backward, and model parameters were updated using the optimizer.

### 4.4 Evaluation of The Classification Results

To evaluate the system's performance, accuracy, precision, recall, and F1 score metrics were used. Table 3 shows the metrics obtained from four different classifications conducted in the study.

Table 3. Metrics obtained as a result of classification of real and synthetic data

| Train/Test Data | Accuracy | Precision | Recall | F1 Score |
|---|---|---|---|---|
| Real/Real | 0.84 | 0.84 | 0.84 | 0.84 |
| Synthetic /Synthetic | 0.99 | 0.98 | 0.98 | 0.98 |
| Real / Synthetic | 0.81 | 0.82 | 0.78 | 0.78 |
| Synthetic / Real | 0.76 | 0.77 | 0.76 | 0.76 |

Firstly, achieving high accuracy, precision, recall, and F1 score values when using real data as both training and test data (Real/Real) for classification indicates that the pre-trained ResNet18 model can be considered as a baseline model for this study. These results reflect the model's ability to classify real-world data accurately.

The lower accuracy, precision, recall, and F1 score values obtained when using real data for training and synthetic data for testing (Real/Synthetic) compared to Real/Real results can be attributed to the fact that the distribution conformity in synthetic data does not perfectly match that of real data, leading to out of distribution data samples in the generated data. These data samples decrease the accuracy rate of the respective classification. However, the similarity between the metric values obtained from Real/Real and Real/Synthetic classifications indicates that synthetic data closely resemble real data.

Achieving nearly 100% accuracy, precision, recall, and F1 score values when using synthetic data as both training and test data (Synthetic/Synthetic) suggests that the model's capacity is sufficient to learn the distribution of the data. These high rates indicate that synthetic data can be easily learned by the model.

The lower metric results obtained from Real/Real classifications compared to Synthetic/Synthetic classifications in both cases of ResNet18 classification can be explained as follows: The real dataset contains out of distribution examples within itself, which negatively affect synthetic data generation and classification results. Upon examination of the dataset, many out of distribution data points were found.

The lower accuracy, precision, recall, and F1 score values obtained when using synthetic data for training and real data for testing (Synthetic/Real) compared to Real/Real classifications by 8% can be interpreted as follows:

- As mentioned earlier, the presence of out of distribution examples within the real dataset and the use of real data as test data affect the classification results. The model could not find correlation for out of distribution data.
- The lesser diversity of synthetic data compared to real data might imply that synthetic data have less diversity than real data (since the area learned from the real data distribution during synthetic data generation is small compared to the total distribution of real data, the diversity within synthetic data is low). When synthetic data is tested with real data, low metric values are obtained due to the low diversity of the training data.

The similar metric values obtained from Synthetic/Real and Real/Synthetic classifications using the ResNet18 model highlight the similarity between real and synthetic data. Furthermore, achieving high results using a dataset containing out of distribution data demonstrates the success of the model.

## 5. CONCLUSION

This article discusses the increasing importance of data in today's world and the growing need for more data in the technological age, despite the current insufficiency of data in certain areas, particularly in healthcare, due to limitations such as data scarcity, data imbalance, data accessibility, and privacy. Therefore, synthetic data is required. Synthetic data can enrich and diversify real datasets. This enables the model to be trained from a broader perspective and adapt to various conditions. As a solution to the problem, this article proposes generating highly realistic synthetic data from breast cancer histopathology images using MSG-GAN and then classifying both real and synthetic data using the pre-trained ResNet18 model to determine the similarity of synthetic data to real data. Additionally, the article discusses algorithms and architectures from ESA and Deep Learning.

This research presents the results of experiments conducted on four different classification tasks to evaluate how close synthetic data is to real data. The results show that synthetic data can perform similarly to real data but must be carefully evaluated. Successful outcomes include the generation of realistic synthetic images, eliminating the need for manual visual breast cancer detection. It is predicted that the production of synthetic data in healthcare will reduce the need for human intervention in the disease detection and diagnosis process and accelerate research in the healthcare field. Therefore, it is concluded that the use of GANs in healthcare for purposes such as data augmentation, dataset enrichment, and balancing is highly beneficial and reliable.

In future studies, the focus should be on improving the process of synthetic data generation using datasets with less skewed distributions and enhancing the quality of synthetic data. Additionally, it is recommended to employ different metrics to further measure the similarity of synthetic data to real data.